\documentclass[letterpaper]{article}
\usepackage{aaai}
\usepackage{times}
\usepackage{helvet}
\usepackage{courier}
\usepackage{url}
\usepackage{booktabs}

\usepackage{pdfsync}

\usepackage{grffile}

\usepackage[ansinew]{inputenc}
\usepackage{graphicx} 
\usepackage{textcomp} 
\usepackage{amsfonts} 
\usepackage{amsmath}
\usepackage{amssymb}
\usepackage{latexsym}
\usepackage{array}
\usepackage{float}
\usepackage{amsthm} 
\usepackage{calc}
\usepackage{perpage}
\usepackage{textcomp}
\usepackage{verbatim} 
\usepackage{marvosym}
\usepackage{hieroglf}
\usepackage{multirow} 
\usepackage{rotating} 
\usepackage{dsfont} 

\usepackage[english]{babel}

\usepackage{tikz}
\usetikzlibrary{shapes,arrows}

\newtheorem{theorem}{Theorem}
\newtheorem{lemma}{Lemma}
\newtheorem{definition}{Definition}

\newcommand{\argmin}{\mathop{\mathrm{arg\,min}}}
\newcommand{\argmax}{\mathop{\mathrm{arg\,max}}}

\def\E{\mathbb{E}}

\def\N{\mathbb{N}}

\def\R{\mathbb{R}}

\newcommand{\mpr}{\mathbb{P}}

\newcommand{\ny}{\color{black}}
\newcommand{\nye}{\color{black}}

\graphicspath{ {./figures/} }

\begin{document}
%

\title{Compute Less to Get More:\\
Using ORC to Improve Sparse Filtering}

\author{Johannes Lederer\\
Department of Statistical Science\\
Cornell University\\
Ithaca, NY 14853\\
johanneslederer@cornell.edu\\
\And
Sergio Guadarrama\\
EECS Department\\
University of California\\
Berkeley, CA 94720\\
sergio.guadarrama@berkeley.edu}

\maketitle

\begin{abstract}
\begin{quote}
Sparse Filtering is a popular feature learning algorithm for image classification pipelines. In this paper, we connect the performance of Sparse Filtering with spectral properties of the corresponding feature matrices. This connection provides new insights into Sparse Filtering; in particular, it suggests early stopping of Sparse Filtering. We therefore introduce the Optimal Roundness Criterion (ORC), a novel stopping criterion for Sparse Filtering. We show that this stopping criterion is related with  pre-processing procedures such as Statistical Whitening and demonstrate that it can make image classification with Sparse Filtering considerably faster and more accurate.
\end{quote}
\end{abstract}

\section{Introduction}

Standard ways to improve image classification are to collect more samples or to change the representation and the processing of the data. In practice, the number of samples is typically limited, so that the second approach becomes relevant. An important tool for this second approach are feature learning algorithms, which aim at easing the classification task by transforming the data. Recently proposed deep learning methods intend to jointly learn  learn a feature transformation and the classification~\cite{Krizhevsky:2012}. In this work, however, we focus on unsupervised feature learning, especially on Sparse Filtering, because of their simplicity and scalability.

Feature learning algorithms for image classification pipelines typically consists of three steps: pre-processing, (un)supervised dictionary learning, and encoding. An abundance of procedures is available for each of these steps, but for accurate image classification, we need procedures that are effective and interact beneficially with each other~\cite{Agarwal06,coates2011importance,coates2011analysis,jia2012beyond,le2013building,LeCun04}. Therefore, a profound understanding of these procedures is crucial to ensure accurate results and efficient computations.

In this paper, we study the performance of Sparse Filtering~\cite{Ngiam:2011} for image classification. Our main contributions are:
\begin{itemize}
\item we show that Sparse Filtering can strongly benefit from early stopping;
\item we show that the performance of Sparse Filtering is correlated with spectral properties of feature matrices on tests sets;
\item we introduce the Optimal Roundness Criterion (ORC), a stopping criterion for Sparse Filtering based on the above correlation, and demonstrate that the ORC can considerably improve image classification.
\end{itemize}

\section{Feature Learning for Image Classification}\label{sec:Feature}

\ny Feature learning algorithms often consist of two steps: In a first step, a dictionary is learned, and in a second step, the samples are encoded based on this dictionary. A typical dictionary learning step for image classification is sketched in Figure~1: First, random patches (samples) are extracted from the training images. These patches are then pre-processed using, for example, Statistical Whitening or Contrast Normalization. Finally, an unsupervised learning algorithm is applied to learn a dictionary from the pre-processed patches. Once a dictionary is learnt, several further steps need to be applied  to finally train an image classifier, see, for example,~\cite{coates2011importance,coates2011analysis,jia2012beyond,le2013building}. Our pipeline is similar to the one in~\cite{coates2011importance}: We extract square patches comprising $9\times 9$ pixels, pre-process them with Contrast Normalization\footnote{Contrast normalization consists of subtracting the mean and dividing by the standard deviation of the pixel values.} and/or Statistical Whitening, and finally pass them to Random Patches or Sparse Filtering. (Note that our outcomes differ slightly from those in~\cite{coates2011importance} because we use square patches comprising $9\times 9$ pixels instead of $6\times 6$ pixels.) Subsequently, we apply soft-thresholding for encoding, $4\times 4$ spatial max pooling for extracting features from the training data images, and finally L2 SVM classification (cf.~\cite{coates2011importance}).\nye

\begin{figure}[h]
  \centering
\scalebox{0.6}{

\tikzstyle{decision} = [diamond, draw, fill=blue!20, 
    text width=4.5em, text badly centered, node distance=3cm, inner sep=0pt]
\tikzstyle{block} = [rectangle, draw, fill=blue!0, 
    text width=6.5em, text centered, rounded corners, minimum height=4em, line width = 2pt]
\tikzstyle{block2} = [rectangle, draw, fill=blue!15, 
    text width=4.5em, text centered, minimum height=1em, line width = 1pt]
\tikzstyle{line} = [draw, -latex']

\begin{tikzpicture}[node distance = 3cm, auto]
\footnotesize     
\node [block2] (training) {Training images};  
 \node [block,right of = training] (extract) {Extraction of random patches\vspace{0.4mm} \includegraphics[width=0.5\textwidth]{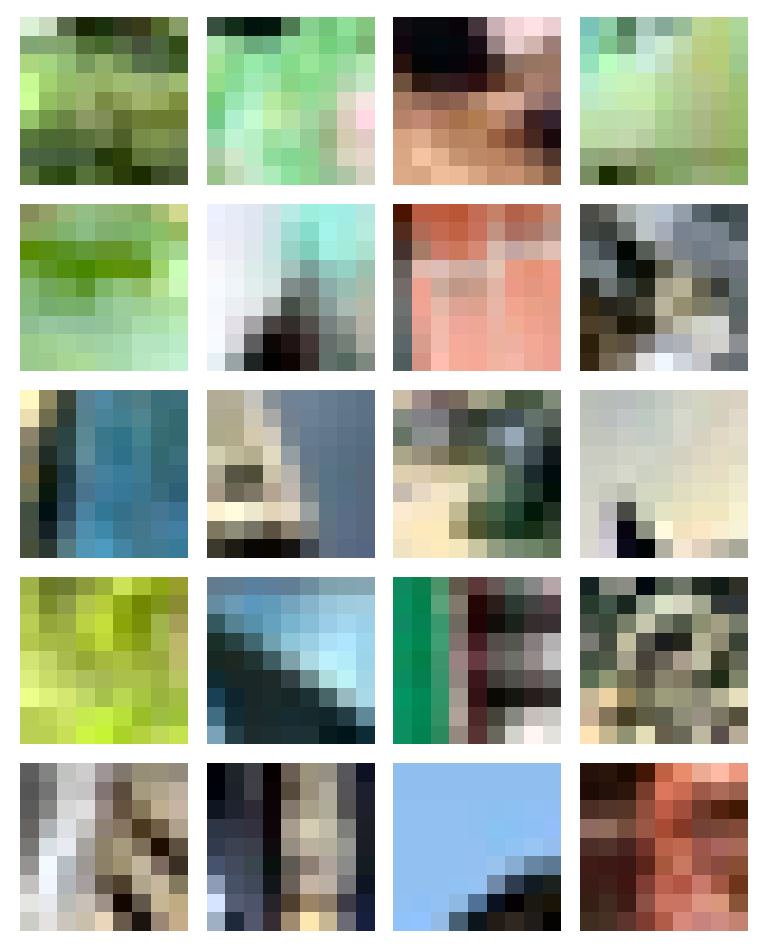}};
 \node [block,right of = extract] (pre) {Pre-processing\vspace{0.5mm} \includegraphics[width=0.5\textwidth]{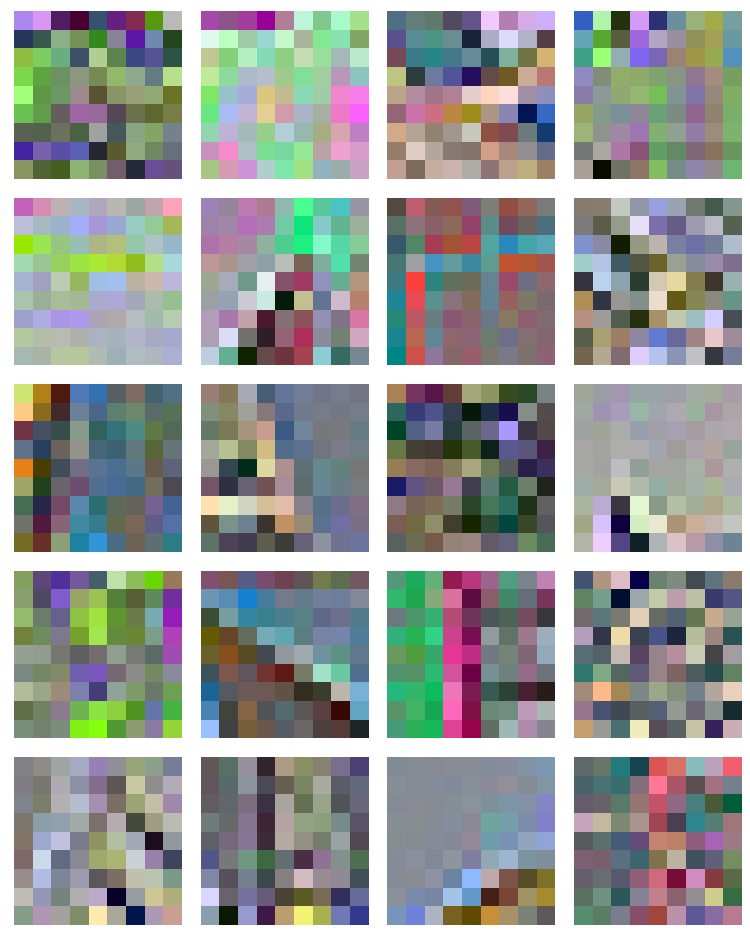}};
 \node [block,right of = pre] (usl) {Unsupervised learning \includegraphics[width=0.5\textwidth]{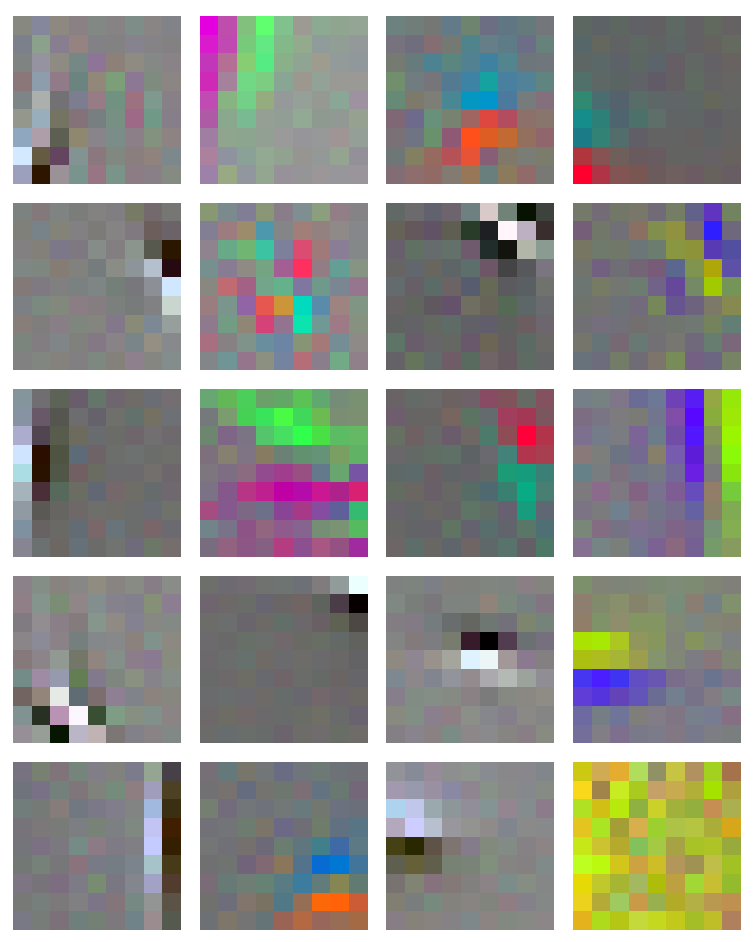}};
 \node [block2,right of = usl] (fe) {Dictionary};

\path [line,dashed] (training) |- (extract);
\path [line] (extract) |- (pre);
\path [line] (pre) |- (usl);
\path [line,dashed] (usl) |- (fe);
\end{tikzpicture}}
\caption{A typical dictionary learning step. Statistical Whitening and Contrast Normalization are examples for pre-processing procedures; Random Patches and Sparse Filtering are examples for unsupervised learning procedures.}
 \label{fig:dict1}
\end{figure}
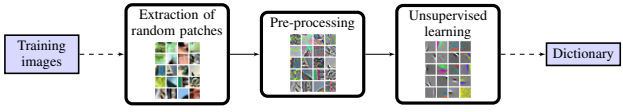

Numerous examples show that feature learning can considerably improve classification. Therefore, insight in the underlying principles of feature learning algorithms such as Statistical Whitening and Sparse Filtering is of great interest.

In mathematical terms, a feature learning algorithm provides a transformation
\begin{equation}\label{d:transformations}
  \begin{aligned}
    \mathcal F:\R^{l\times p}&\to\R^{n\times p}\\
X&\mapsto \mathcal F(X)
  \end{aligned}
\end{equation}
of an original feature matrix $X\in\R^{l\times p}$ to a new feature matrix $\mathcal F(X)\in\R^{n\times p}$. We adopt the convention that the rows of the matrices correspond to the features, the columns to the samples; this convention implies  in particular that $l\in\N$ is the number of original features, $p\in\N$ the number of samples, and $n\in\N$ the number of new features. 

\section{The Optimal Roundness Criterion}\label{sec:ORC}

\subsection{Roundness of Feature Matrices}

\ny Feature learning can be seen as trade-off between reducing the correlations of the feature representation and preservation of relevant information. This trade-off can be readily understood looking at Statistical Whitening. For this, recall that pre-processing with Statistical Whitening  transforms a set of image patches into a new set of patches by changing the local correlation structure. More precisely, Statistical Whitening transforms patches~$X_{\text{Patch}}\in\R^{n'\times p}$ ($n'<n$), that is, subsets of the entire feature matrix, into new patches $\mathcal F_{\text{Patch}}(X_{\text{Patch}})$ such that
\begin{align*}
  \mathcal F_{\text{Patch}}(X_{\text{Patch}})^T\mathcal F_{\text{Patch}}(X_{\text{Patch}})=n'\operatorname{I}_{n'}.
\end{align*}
Statistical Whitening therefore acts locally: while the correlation structures of the single patches are directly and radically changed, the structure of the entire matrix is affected only indirectly. However, these indirect effects on the entire matrix are important for the following. \nye To capture these effects, we therefore introduce the roundness of a feature matrix~$F:=\mathcal F(X)$ given an original feature matrix~$X$. On a high level, we say that the new feature matrix $F$ is round if the spectrum of the associated Gram matrix $FF^T\in \R^{n\times n}$ is narrow. To specify this notion, we denote the ordered eigenvalues of $FF^T$ by $\sigma_1(F)\geq \dots \geq \sigma_n(F)\geq 0$ and their mean by  $\overline\sigma(F):=\frac{1}{n}\sum_{i=1}^n\sigma_i(F)$ and define roundness as follows:
\begin{definition}\label{d:roundness} For any  matrix $F\neq 0$, we define its roundness as  
  \begin{equation*}
    \label{eq:max}
    r(F):=\frac{\overline\sigma(F)}{\sigma_1(F)}\in [0,1].
  \end{equation*}
\end{definition}
\noindent The largest eigenvalue $\sigma_1$ measures the width of the spectrum of the Gram matrix; alternative measures of the width such as the standard deviation of the eigenvalues would serve the same purpose. The mean of the eigenvalues $\overline\sigma$, on the other hand, is basically a normalization as the following result illustrates (the proof is found in the supplementary material):  
\begin{theorem}\label{r:mean} Denote the columns of $F$ by $F^1,\dots,F^p$. Then, the mean $\overline\sigma(F)$ of the eigenvalues of the Gram matrix $FF^T$ is constant on $\mathcal S:=\{ F\in \R^{n\times p}: \|F^1\|_2=\dots=\|F^p\|_2= 1\}:$
\begin{equation*}
  \label{eq:mean}
  \overline\sigma(F):= \frac p n\text{~~~~for all~}F\in\mathcal S. 
\end{equation*}
\end{theorem}
\noindent Definition~1 therefore states that the larger is $r$, the narrower is the spectrum of the eigenvalues of the Gram matrix of $F$, and therefore, the rounder is the matrix $F$. \ny With this notion of roundness at hand, we can now understand the effects of Statistical Whitening: On the one hand, Definition~1 indicates that Statistical Whitening renders single patches perfectly round, that is, $r(\mathcal F_{\text{Patch}})=1$. On the other hand, Statistical Whitening preserves global structures in the feature matrix. In particular, the entire feature matrix is made rounder but {\it not} rendered perfectly round, that is, $r(\mathcal F (X))<1$. In this sense, Statistical Whitening can be seen as trade-off between increasing of roundness and preservation of global structures. It therefore remains to connect roundness and randomization.\nye

\subsection{Roundness and Randomness}

\ny A connection between roundness and randomization is provided by random matrix theory. \nye  To illustrate this connection, we first recall Gordon's theorem for Gaussian random matrices (see~\cite[Chapter 5]{Eldar:2012} for a recent introduction to random matrix theory):
\begin{theorem}[Gordon]\label{thm:gordon}
  Let $F\in\R^{n\times p}$ be a random matrix with independent standard normal entries. Then, 
  \begin{align*}
&1-\sqrt{{n}/{p}}\leq    \E\left[\sqrt{ \sigma_n(F)/p}\right]\\
&~~~~~~~~~~~~~~~~~~~~~~~~~\leq \E\left[\sqrt{\sigma_1(F)/p}\right]\leq 1+\sqrt{{n}/{p}}.
  \end{align*}
\end{theorem}
\noindent Such exact bounds are available only for matrices with independent standard normal entries, but sharp bounds in probability are available also for other random matrices. For our purposes, the common message of all these bounds is that random matrices with sufficiently many columns (number of samples) have a small spectrum. This means in particular that such matrices are round as the following asymptotic result illustrates (the proof is based on well-known results from random matrix theory and therefore omitted):
\begin{lemma}\label{lemma:gordon}
Let the number of features $n\equiv n(p)$ be a function of the number of samples $p$ such that $n/p\to 0$. Moreover, for all $p\in\{1,2,\dots\}$, let  $F\equiv F(p)$ be a random matrix with independent standard normal entries. Then, for all $\epsilon>0$,
  \begin{equation*}
  \mpr\left(|r(F)-1|> \epsilon\right)\to 0 \text{~~~~for}~~p\to\infty,  
  \end{equation*}
that is, $r(F)$ converges in probability to $1$.
\end{lemma}
\noindent  Similar results can be derived for non-Gaussian or correlated entries, indicating that random matrices are typically round.\nye

\ny Besides the connection between roundness and randomization, the above results for random matrices also provide a link between roundness and sample sizes. Indeed, we observe that the above results indicate that large samples sizes lead to round matrices. To make this link more tangible,  \nye we conduct simulations with Toeplitz matrices, which can model local correlations that are typical for nearby pixels in natural images \cite{girshick2013training}.  To this end, we first recall that for any fixed parameter~$\rho\in [0,1)$, the entries of a Toeplitz matrix $T_\rho$  are defined as\footnote{We set $0^{k}:=1$ for $k=0$ and $0^{k}:=0$ for $k\neq 0$.} $ (T_\rho)_{ij}:=\rho^{|i-j|}$. We now construct a feature matrix $U_{\rho}$ by drawing each of its columns, that is, the samples, from the normal distribution with mean zero and covariance matrix $T_\rho\in\R^{n\times n}$. Toeplitz matrices with $\rho=0$ lead to feature matrices with independent entries; Toeplitz matrices with $\rho=0.8$ lead to feature matrices with dependence structures that are more similar to dependence structures found in natural images. In Figure~2, we report the roundness of $U_{\rho}$ for $\rho=0$ (plot on the left) and $\rho=0.8$ (plot on the right) as a function of the numbers of samples~$p$ for different numbers of features~$n$.  The results are commensurate with the theoretical findings above: First, both plots illustrate that the roundness of matrices increases if the number of samples is increased but decreases if the number of features is increased (cf.~Theorem~2 and Lemma~1). Second, a comparison of the two plots illustrate that the roundness is larger for $\rho=0$ than for $\rho=0.8$ (since $T_0$ is perfectly round while $T_{0.8}$ is not).

\begin{figure}[h]
  \centering
\includegraphics[width=0.22\textwidth]{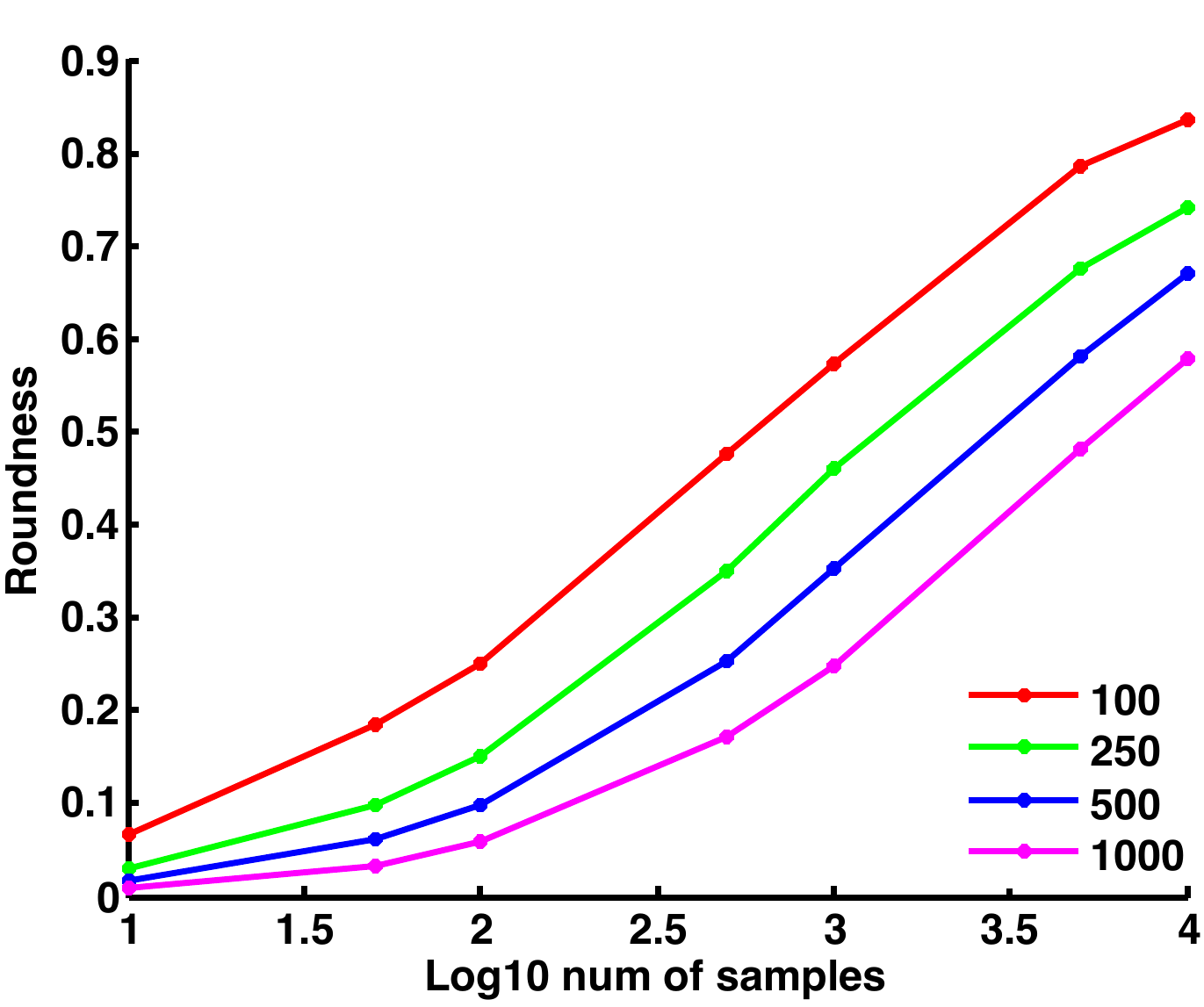}~~~~~~\includegraphics[width=0.22\textwidth]{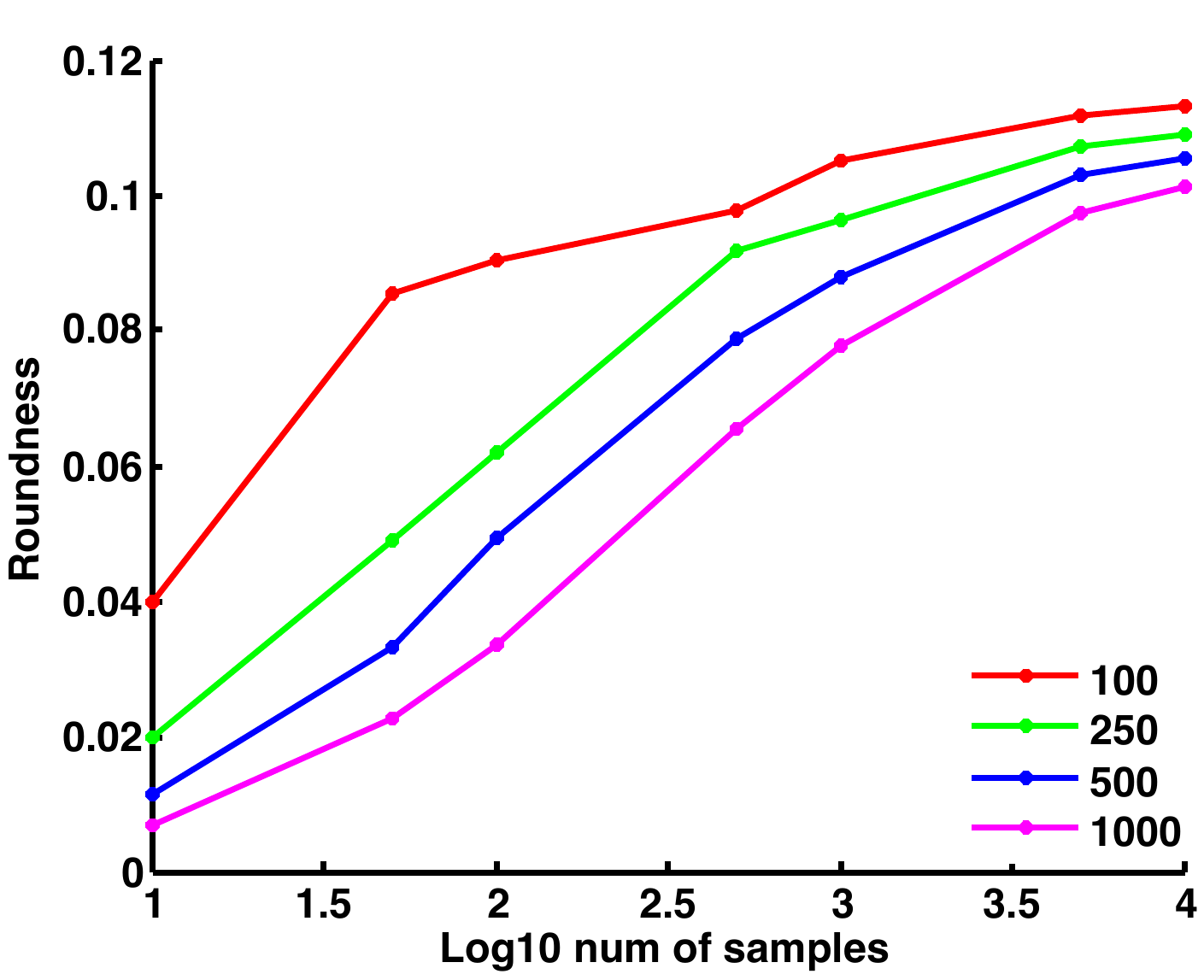}  
  \caption{Roundness of random feature matrices with columns drawn from a normal distribution with mean zero and a Toeplitz covariance matrix. For the left plot,  $\rho=0$ (uncorrelated entries). For the right plot, $\rho=0.8$ (correlations that relate to natural images). The roundness is plotted as a function of the number of samples $p\in[10,10^4]$ for four different numbers of features $n\in\{100, 250, 500, 1000\}$.}
  \label{fig:toep}
\end{figure}

\subsection{Optimal Roundness Criterion (ORC)}

\ny The above discussion suggest that optimal feature learning is the result of a trade-off between increasing the roundness of the feature matrix and preserving global structures in the data. In this part, we want to exploit this insight to understand and improve iterative feature learning algorithms. \nye Common feature learning algorithms consist of transformations that are defined as minimizers of a functional. These functionals are then often computed iteratively via a sequence of gradient based operations. In this paper, we therefore focus on feature learning algorithms where the transformation $\mathcal F$ as in~(1) is the limit of a sequence of transformations $(\mathcal G_k)_{k\in\N}$, that is,
\begin{equation*}
  \mathcal F = \lim_{k\to \infty}\mathcal G_k,
\end{equation*}
where for all $k\in\N$,
\begin{align*}
  \mathcal G_k:\R^{l\times p}&\to\R^{n\times p}\\
X&\mapsto \mathcal G_k(X).
\end{align*}
\ny A prominent representative of such iterative algorithms is Sparse Filtering. Sparse Filtering consists of normalizations and the minimization of an $\ell_1$-criterion (see next Section). It is reasonable to assume that these operations - similar to the local changes by Statistical Whitening - preserve certain global structures of the feature matrix. In view of a trade-off between roundness and preservation of global structures, we are therefore interested in stopping the iterations as soon as the roundness is maximized. More formally, we introduce the ORC, which serves as stopping criterion to maximize the roundness:\nye
\begin{definition}\label{d:stopping}
Let $r$ be the roundness introduced in Definition~1. The Optimal Roundness Criterion (ORC) replaces the transformation $\mathcal F$ by 
\begin{equation*}
    \widehat{\mathcal{G}}:= \mathcal G_{\widehat k}
\end{equation*}
for 
\begin{equation*}
  \widehat k := \argmax_{k\in\N}\{r(\mathcal G_{k'})<r(\mathcal G_{k})\text{~for all~} k'<k\}
\end{equation*}
if the arg-maximum is finite and  $\widehat{\mathcal G}:=\mathcal F$ otherwise.
\end{definition}
\noindent  The ORC assures that the computations continue only as long as the roundness increases. Assuming that certain global structures are preserved by the transformations, the ORC provides an optimization scheme for the performance of iterative feature learning algorithms. \nye One could also think of modifications of the ORC that include an additive constant or a factor to force larger increases or to allow for temporary decreases of the roundness.

\section{Image Classification on CIFAR-10}\label{sec:applications}

For our all experiments, we use the CIFAR-10 dataset \cite{krizhevsky2009learning}\footnote{\url{http://www.cs.toronto.edu/~kriz/cifar.html}}. This dataset consists of $60\ 000$ color images partitioned into $10$ classes, each containing $6\ 000$ images. Each of the images comprises $32\times 32$ pixels. The dataset is split into a training set with $50\ 000$ images and a test set with $10\ 000$ images. From the training set, we randomly select $10\ 000$ patches for the unsupervised feature learning. These patches are also used to determine the parameters of Contrast Normalization and Statistical Whitening (if applied).

\subsection{Random Patches}
For the dictionary learning step, it was shown that simple randomized procedures combined with Statistical Whitening work surprisingly well \cite{coates2011importance,Jarrett09,Saxe11}. A popular example is Random Patches, which creates a dictionary matrix by simply stacking up randomly selected samples. In Table~1, we report the influence of Contrast Normalization and Statistical Whitening on Random Patches (cf.~\cite{coates2011importance}). We see that Statistical Whitening is very beneficial for Random Patches and increases the roundness of the transformed feature matrix. This suggests that the roundness can be used as an indicator for the performance of feature learning. (Note that the roundness is on different scales for different numbers of features and can therefore not be compared for different numbers of features.)

\begin{table}[!htb]
\begin{center}
  \begin{scriptsize}
\begin{tabular}{ r | c c|  c c}
\toprule
Num & Norm. & White. & Round.  & Acc. \\
\midrule
243 & No & No & 0.0041 & 32.50\% \\
243 & Yes & No & 0.0131 & 63.65\% \\
243 & No & Yes & 0.2080 & 65.01\% \\
243 & Yes & Yes & 0.1548 & 64.34\% \\
\midrule
486 & No & No & 0.0021 & 31.67\% \\
486 & Yes & No & 0.0062 & 66.14\% \\
486 & No & Yes & 0.1134 & 67.08\% \\
486 & Yes & Yes & 0.0965 & 67.84\% \\
\bottomrule
\end{tabular}
\end{scriptsize}
\end{center}
\caption{Roundness and accuracy of Random Patches with and without Contrast Normalization and Statistical Whitening.}
\label{table:random_patches}
\end{table}

\begin{table}[h]
\begin{center}
\begin{scriptsize}
\begin{tabular}{ r | c|  c c}
\toprule
Num & ORC & Round.  & Acc. \\
\midrule
243 & No & 0.0519 & 57.66\%\\
243 & Yes & 0.1425 & 62.47\%\\
\midrule
486 & No & 0.0495 & 58.19\%\\
486 & Yes & 0.0908 & 63.80\%\\
\bottomrule
\end{tabular}
\end{scriptsize}
\end{center}
\caption{Roundness and accuracy of Sparse Filtering with and without early stopping based on the ORC.}
\label{table:sf}
\end{table}


\subsection{Sparse Filtering}\label{sec:SF}

Sparse Filtering~\cite{Ngiam:2011} is an unsupervised feature learning algorithm that computationally scales particularly well with the dimensions. To recall the definition of Sparse Filtering, we denote by $\mathcal N:\R^{n\times p}\to \R^{n\times p}$  the function that first normalizes\footnote{We set $0/0:=0*\infty:=\infty$ in the corresponding operations ensure that (2) is well defined.} the rows of a matrix in $\R^{n\times p}$ to unit Euclidean norm and then normalizes the columns of the resulting matrix to unit Euclidean norm. For any fixed matrix $X\in\R^{l\times p}$, we then define a matrix $W_X\in\R^{n\times l}$ such that
\begin{equation}\label{d:W}
  W_X\in\argmin_{W\in\R^{n\times l}}\|\mathcal N(WX)\|_1
\end{equation}
if the minimum is finite and $W_X:=0$ otherwise. Sparse Filtering is then the transformation 
\begin{align}\tag{Sparse Filtering}
  \begin{aligned}
  \mathcal{F}_{SF}:\R^{l\times p}&\to\R^{n\times p}\\
X&\mapsto \mathcal F_{SF}(X):=W_XX.    
  \end{aligned}
\end{align}
However, we now show by making the normalizations explicit that these normalizations make Sparse Filtering intricate.  For this, we define the rank one matrices $E_1,\dots,E_n\in\R^{n\times n}$ via
\begin{equation*}
  (E_i)_{kl}:=\delta_{kl}\delta_{ik}\qquad\forall i,k,l\in\{1,\dots,n\}
\end{equation*}
and $G_1,\dots,G_p\in\R^{p\times p}$ via
\begin{equation*}
  (G_i)_{kl}:=\delta_{kl}\delta_{ik}\qquad\forall i,k,l\in\{1,\dots,p\}
\end{equation*}
where $\delta$ is the usual Kronecker delta. This then yields the following form of Definition~(2).
\begin{theorem}\label{r:sfa}
 The matrix $W_X$ in~(2) is the minimizer of 
{\small
\begin{align*}
&\Bigg \|\left[\sum_{i=1}^nE_iWX(WX)^TE_i\right]^{\text -\frac 1 2}WX~\times\\
&\times~\left[\sum_{i=1}^pG_i (WX)^T\left[\sum_{i=1}^nE_iWX(WX)^TE_i\right]^{\text -1}WXG_i\right]^{\text -\frac 1 2}\Bigg\|_1
\end{align*}}
over all matrices $W\in\R^{l\times n}$.
\end{theorem}
\noindent Although Sparse Filtering is sometimes claimed to have sparsity properties due to the involvement of the $\ell_1$-norm (similar as the Lasso~\cite{Tibshirani-LASSO}, for example), the above reformulation demonstrates that this is far from obvious and needs further clarification.

\begin{figure}[t]
  \centering
  \includegraphics[width=0.4\textwidth]{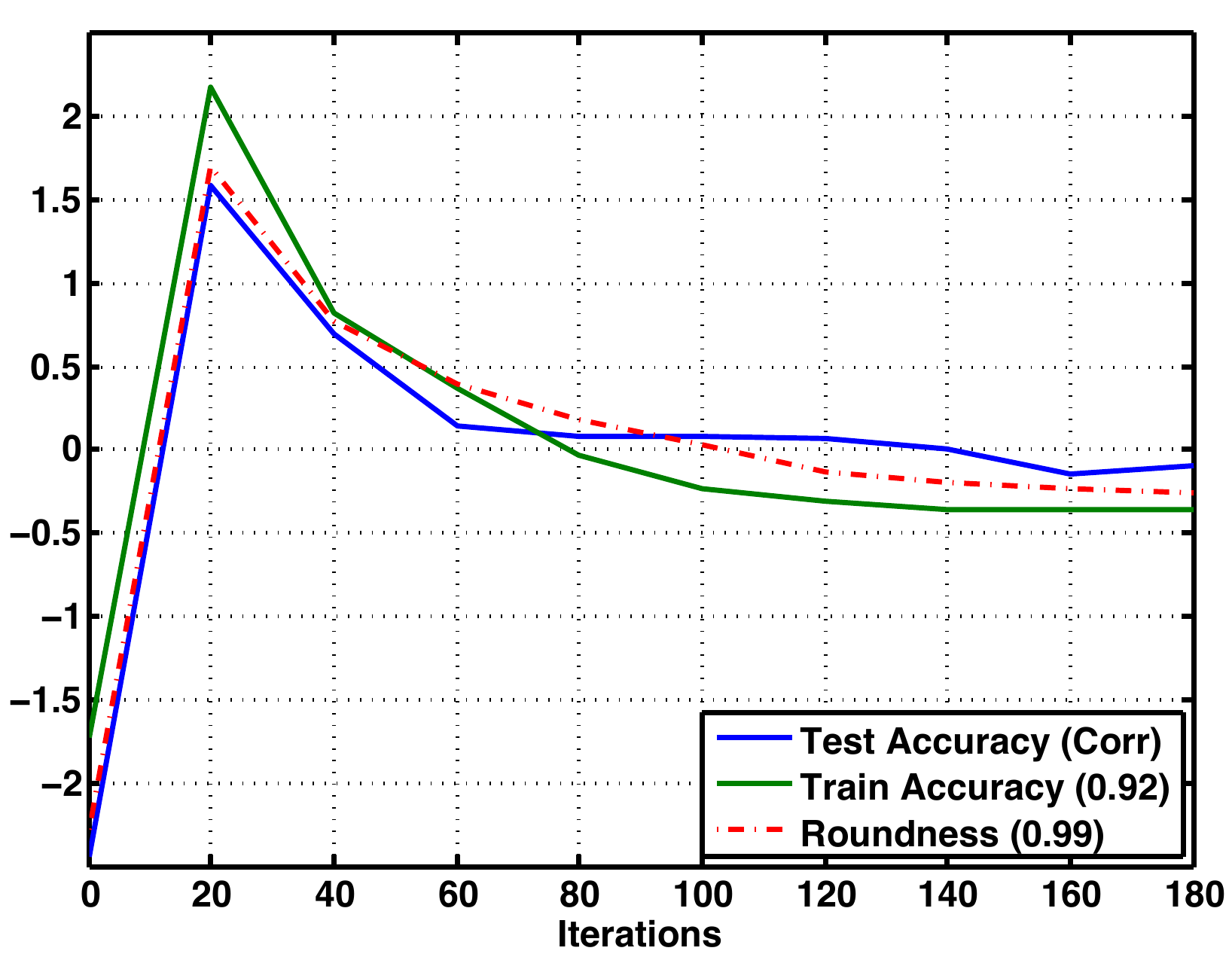}
\vspace{-1.5mm}  \caption{Intermediate outcomes of Sparse Filtering at every 20 iterations. Test accuracy (blue, solid curve), training accuracy (green, solid curve), roundness on the training set (red, dashed curve), and correlations with the test accuracy (numbers in brackets). To enhance visibility, all curves are normalized to have zero mean and unit standard deviation.}
  \label{fig:asym}
\end{figure}  

\begin{figure}[!htb]
  \centering
\includegraphics[width=.4\textwidth]{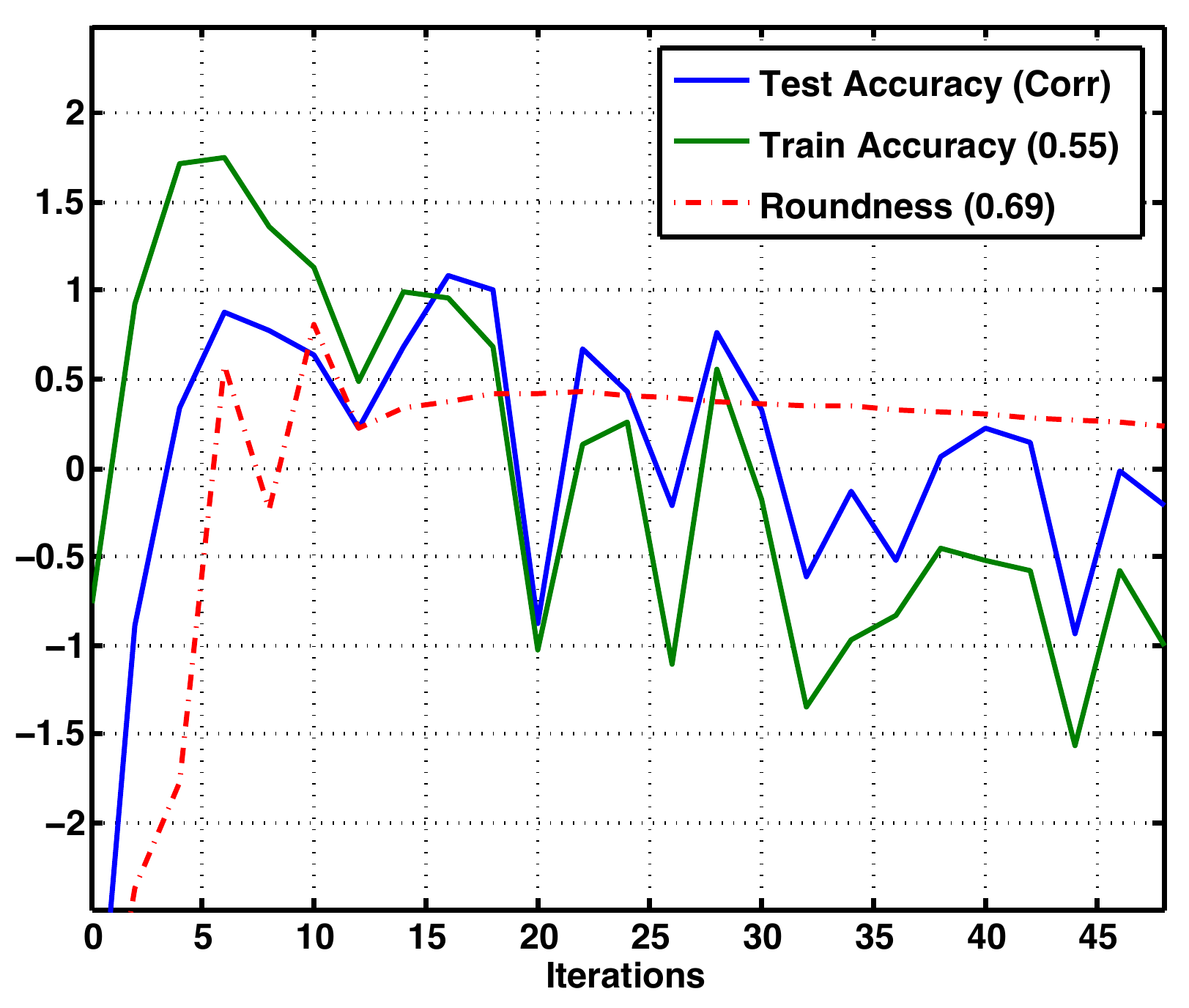} 
\vspace{-1.5mm}
  \caption{Intermediate outcomes of Sparse Filtering at every 2 iterations. Test accuracy (blue, solid curve), training accuracy (green, solid curve), roundness on the training set (red, dashed curve), and correlations with the test accuracy (numbers in brackets). To enhance visibility, all curves are normalized to have zero mean and unit standard deviation.}
  \label{fig:detail1}
\end{figure}

It is apparent that the choice of the number of features~$n$ influences the performance of Sparse Filtering. As can be seen in Figure~/ref{We will see below, however, that the choice of the number of iterations surprisingly can  have an even larger influence. We are therefore interested in choosing an appropriate number of iterations. A standard approach would involve $l$-fold cross-validation schemes, but this requires training of $l$ models and is therefore computationally costly. The ORC, on the other hand, can be a computationally feasible alternative to cross-validation. To illustrate this, we compare  in Table~2 the outcomes of Sparse Filtering on the CIFAR-10 dataset with and without application of the ORC. We have also computed the intermediate  outcomes of Sparse Filtering at every 20~iterations and report in Figure~3 the corresponding test accuracy, training accuracy, roundness on the training set, and correlations with the test accuracy. The roundness on the test set is basically indistinguishable from the roundness on the training set and is therefore not shown. We make three crucial observations: {\it (i)}~the test accuracy of Sparse Filtering peaks at around 20~iterations and then decreases monotonically; {\it (ii)}~the roundness on the training set is highly correlated with the test accuracy; in particular, the locations of the peaks of these curves coincide; {\it (iii)}~the roundness on the training set is highly correlated with the roundness on the test set. These observations suggest that {\it (i)}~Sparse Filtering should be stopped early; {\it (ii)} the ORC can optimize the performance of Sparse Filtering; {\it (iii)} it is sufficient to compute the roundness on the training set. To further support these claims, we have also computed the intermediate outcomes of Sparse Filtering at every 2~iterations in  the region around the peaks, that is, we have computed a zoomed-in version of Figure~3. We report the results in Figure~4. We observe that training accuracy, test accuracy, and roundness are highly correlated, which corroborates the above claims and therefore confirms the potential of the ORC. We finally note that the curves in the zoomed-in version are wiggly not only because of the randomness involved but also because computations of gradients over a small number of iterations involve numerical imprecisions. 

\section{Conclusions and Outlook}\label{sec:discussion}
The spectral analysis of feature matrices is a novel and promising approach to feature learning. In particular, our results show that this ``geometric'' approach can provide new interpretations and substantial improvements of wide-spread feature learning tools such as Statistical Whitening, Random Patches, and Sparse Filtering. For example, we have revealed that Sparse Filtering can, quite surprisingly, deteriorate with increasing number of iterations and can be made considerably faster and more accurate by early stopping according to the spectrum of the intermediate feature matrices. 

Regarding the theory, it would be of interest to obtain, for specific procedures, predictions on how the roundness changes with the iterations and to what it converges in the limit.

In an extended version of this paper, we are planning to include an analysis of Roundness in Convolutional Neural Networks~(CNNs)~\cite{Fukushima:80}. After being neglected for many years, CNNs have received an enormous deal of attention recently, see~\cite{Krizhevsky:2012,Girshick:13} and many others. We therefore expect that the application of our approach to CNNs can be of substantial interest.


\section*{Acknowledgments}
We thank the reviewers for their insightful comments.

\section{Appendix: Proofs}

We denote in the following the columns and rows of any matrix $M$ by $M^1,\dots,M^p$ and $M_1,\dots,M_n$, respectively.
\begin{proof}[Proof of Theorem~1]
  The matrix $FF^T$ is symmetric and can therefore be diagonalized. This implies that there is an orthogonal orthogonal matrix $A\in\R^{n\times n}$ such that the diagonal entries of $AFF^TA^T\in\R^{n\times n}$ are  $\sigma_1(F),\dots, \sigma_n(F)$. For this matrix $A$, it then holds 
  \begin{align*}
\overline\sigma(F)=&    \frac{1}{n}\sum_{i=1}^n\sigma_i(F)\\
=&\frac 1 n \sum_{i=1}^n(AFF^TA^T)_{ii}\\
=&\frac 1 n \operatorname{trace}(AFF^TA^T).
  \end{align*}
Next, we invoke the cyclic property of the trace and the orthogonality of the matrix~$A$ to obtain 
  \begin{align*}
    \operatorname{trace}(AFF^TA^T)=&\operatorname{trace}(F^TA^TAF)\\
=&\operatorname{trace}(F^TF)\\
=&\sum_{i=1}^p(F^TF)_{ii}.
  \end{align*}
Finally, we note that the normalization of the columns of $F$ yields
\begin{equation*}
  (F^TF)_{ii}=(F^i)^TF^i=\|F^i\|_2^2=1
\end{equation*}
for all $i\in\{1,\dots, p\}.$ The desired result
\begin{equation*}
   \overline\sigma(F)=\frac{1}{n}\sum_{i=1}^p 1=\frac p n
\end{equation*}
can now derived combining the three displays.
\end{proof}

\begin{proof}[Proof of Theorem~3] We first show that for a matrix $A\in\R^{n\times p}$, the corresponding matrix $A^R\in\R^{n\times p}$ with normalized rows can be written as
\begin{equation*}
  \tag{Claim 1}A^R=\left[\sum_{i=1}^nE_iAA^TE_i\right]^{\text -1/2}A.
\end{equation*}
To this end, we observe that the normalization of the rows of the matrix $A$ corresponds to the matrix multiplication
\begin{equation*}
  A^R=D^RA,
\end{equation*}
where $D^R\in\R^{n\times n}$ is the diagonal matrix with nonzero entries
\begin{equation*}
  (D^R)_{ii}=1/\sqrt{\sum_{j=1}^p(A_{ij})^2}\qquad\forall i\in\{1,\dots,n\}.
\end{equation*}
Next, we note that 
\begin{equation*}
  \sum_{j=1}^p(A_{ij})^2=(AA^T)_{ii}\qquad\forall i\in\{1,\dots,n\}
\end{equation*}
and therefore
\begin{equation*}
  ((D^R)^{\text -1})_{ii}=1/(D^R)_{ii}=\sqrt{(AA^T)_{ii}}\qquad\forall i\in\{1,\dots,n\}.
\end{equation*}
This yields the matrix equation
\begin{equation*}
  (D^R)^{\text -2}=\sum_{i=1}^nE_iAA^TE_i
\end{equation*}
and therefore
\begin{equation*}
  D^R=\left[\sum_{i=1}^nE_iAA^TE_i\right]^{\text -1/2}.
\end{equation*}
This proves the first claim.\\
We now show that for a matrix $B\in\R^{n\times p}$, the corresponding matrix $B^C\in\R^{n\times p}$ with normalized columns is given by
\begin{equation*}
  \tag{Claim 2}B^C=B\left[\sum_{i=1}^pG_iB^TBG_i\right]^{\text -1/2}.
\end{equation*}
We first note that we can write the normalization step - this time for the columns - as the matrix multiplication
\begin{equation*}
  B^C=B D^C,
\end{equation*}
where $D^C\in\R^{p\times p}$ is the diagonal matrix with entries
\begin{equation*}
  (D^C)_{ii}=1/\sqrt{\sum_{j=1}^n(B_{ji})^2}\qquad\forall i\in\{1,\dots,p\}.
\end{equation*}
Next, we note that 
\begin{equation*}
  \sum_{j=1}^n(B_{ji})^2=(B^TB)_{ii}\qquad\forall i\in\{1,\dots,p\},
\end{equation*}
and therefore for the inverse diagonal matrix
\begin{equation*}
  ((D^C)^{\text -1})_{ii}=1/(D^C)_{ii}=\sqrt{(B^TB)_{ii}}\qquad\forall i\in\{1,\dots,p\}.
\end{equation*}
This yields the matrix equation
\begin{equation*}
  (D^C)^{\text -2}=\sum_{i=1}^pG_iB^TBG_i
\end{equation*}
and therefore
\begin{equation*}
  D^C=\left[\sum_{i=1}^pG_iB^TBG_i\right]^{\text -1/2}.
\end{equation*}
This proves the second claim.\\
We now consider $F_W:=WX\in\R^{n\times p}$ for an arbitrary matrix $W\in\R^{n\times p}$ and apply Claim 1 and Claim 2: Setting $A=F_W$, we obtain from Claim 1 that normalizing the rows of the  matrix $F_W$ yields the matrix $F_W^R\in\R^{n\times p}$ given by
\begin{equation*}
  F^R_W:=\left[\sum_{i=1}^nE_iF_WF_W^TE_i\right]^{\text -1/2}F_W.
\end{equation*}
This implies in particular
\begin{align*}
  (F_W^R)^TF_W^R=&F_W^T\left[\sum_{i=1}^nE_iF_WF_W^TE_i\right]^{\text -1}F_W.
\end{align*}
Setting then $B=F^R_W$, we obtain from Claim 2 and the two previous displays that the matrix $F_W$ becomes after normalizing its rows and then its columns the matrix
\begin{align*}
&  \left[\sum_{i=1}^nE_iF_WF_W^TE_i\right]^{\text -1/2}F_W~\times\\
&~~~~\times~\left[\sum_{i=1}^pG_i F_W^T\left[\sum_{i=1}^nE_iF_WF_W^TE_i\right]^{\text -1}F_WG_i\right]^{\text -1/2}.
\end{align*}
The desired result can then be deduced from the definition of $W_X$ in~(2).
\end{proof}

\appendix
\subsection{Appendix}

We also present numerical outcomes for a different random splitting of the CIFAR-10 dataset. In particular, we recompute Figures~\ref{fig:asym} and~\ref{fig:detail1} for a different splitting and give the results in Figures~\ref{fig:asymappendix} and~\ref{fig:detail1appendix} below. The conclusions are virtually the same as above, which further corroborates our findings.

\begin{figure}[t]
  \centering
  \includegraphics[width=0.4\textwidth]{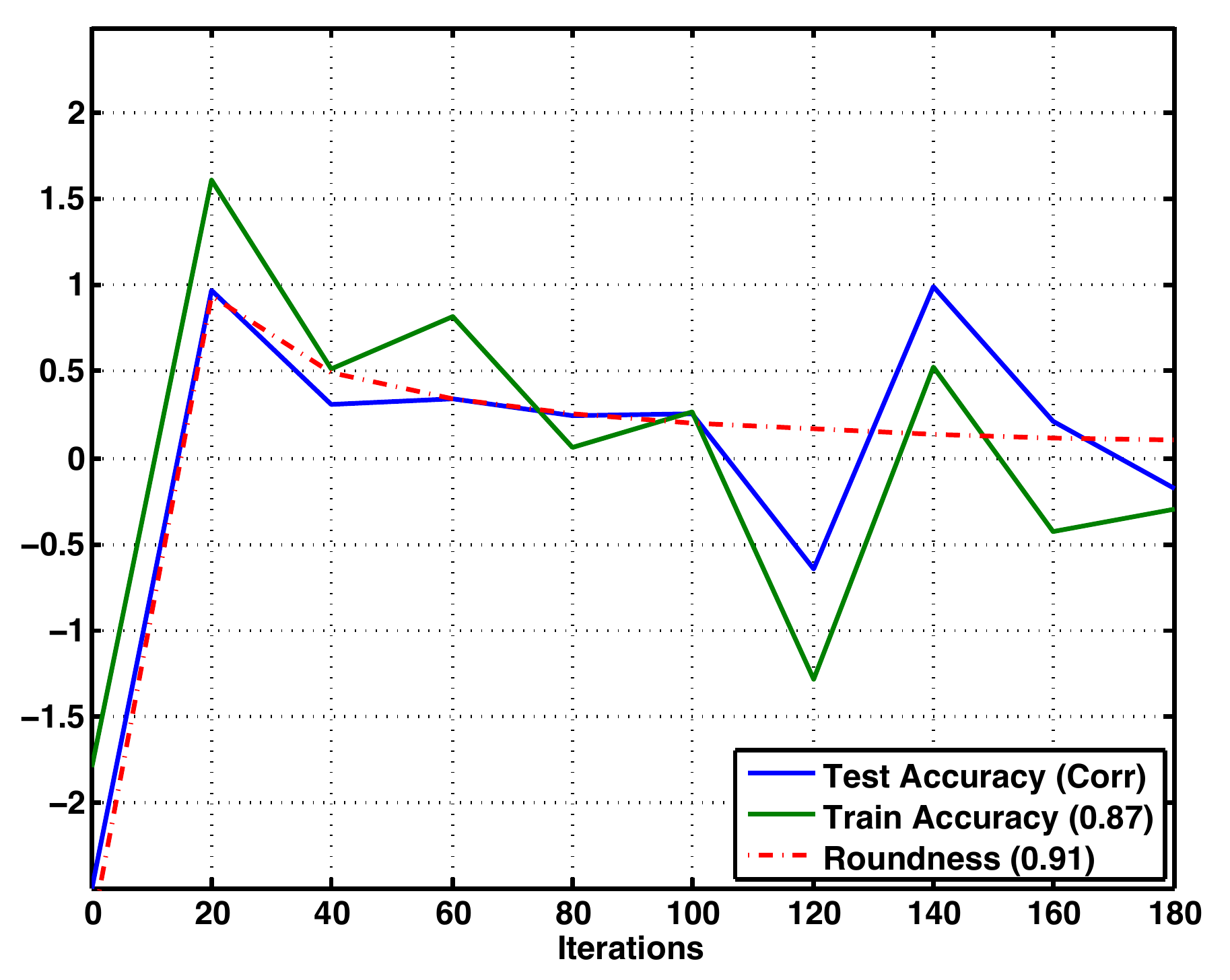}
\vspace{-1.5mm}  \caption{Intermediate outcomes of Sparse Filtering at every 20 iterations. Test accuracy (blue, solid curve), training accuracy (green, solid curve), roundness on the training set (red, dashed curve), and correlations with the test accuracy (numbers in brackets). To enhance visibility, all curves are normalized to have zero mean and unit standard deviation.}
  \label{fig:asymappendix}
\end{figure}  

\begin{figure}[!htb]
  \centering
\includegraphics[width=.4\textwidth]{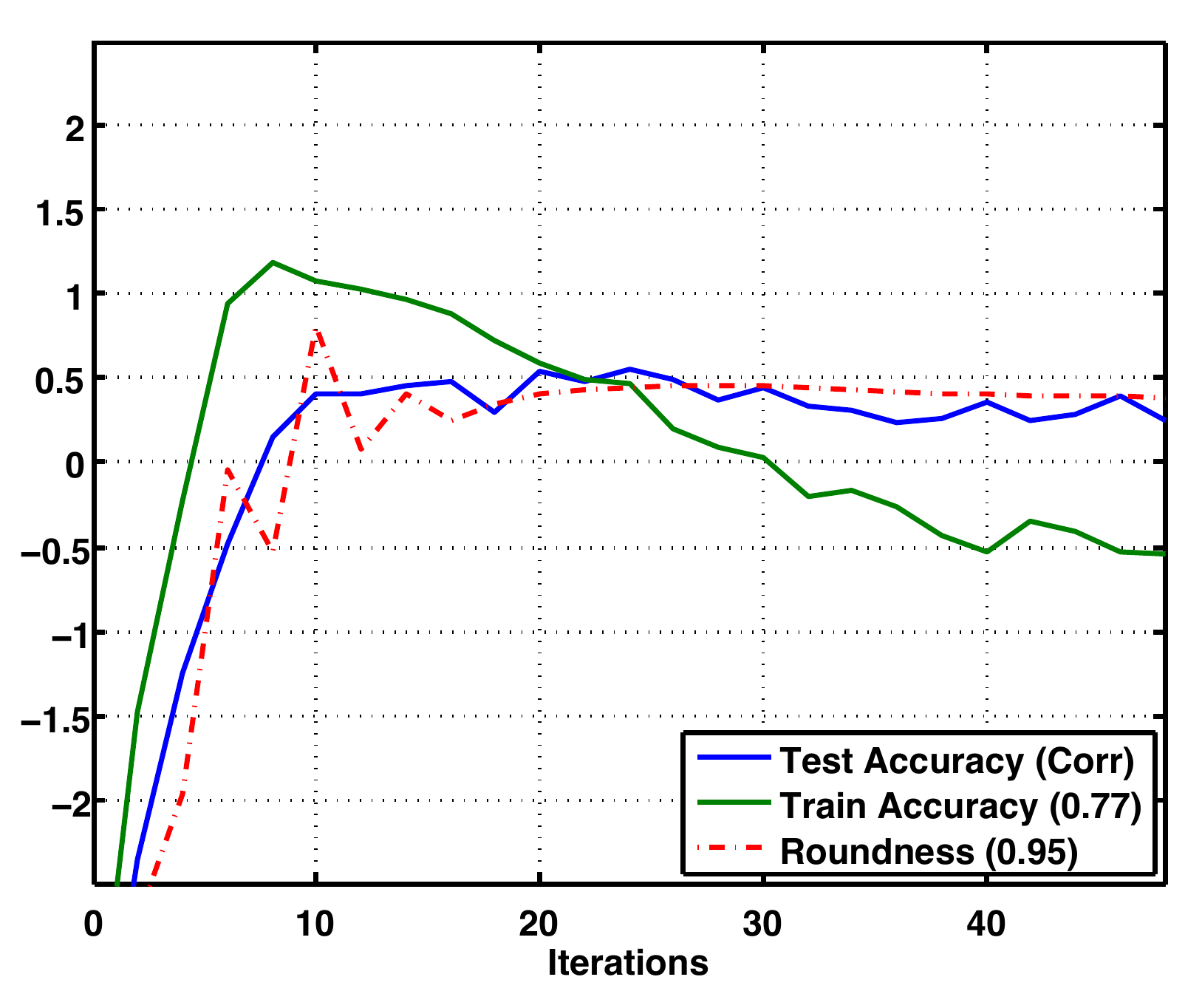} 
\vspace{-1.5mm}
  \caption{Intermediate outcomes of Sparse Filtering at every 2 iterations. Test accuracy (blue, solid curve), training accuracy (green, solid curve), roundness on the training set (red, dashed curve), and correlations with the test accuracy (numbers in brackets). To enhance visibility, all curves are normalized to have zero mean and unit standard deviation.}
  \label{fig:detail1appendix}
\end{figure}  

\bibliographystyle{aaai}
\bibliography{Literature}
\end{document}